# CORD19STS: COVID-19 Semantic Textual Similarity Dataset

Xiao Guo*, Hengameh Mirzaalian*, Ekraam Sabir,

Ayush Jaiswal *and* Wael Abd-Almageed

USC Information Sciences Institute, Marina Del Rey, CA, USA

`{xiaoguo, hengameh, esabir, ajaiswal, wamageed}@isi.edu`

**Abstract**—In order to combat the COVID-19 pandemic, society can benefit from various natural language processing applications, such as dialog medical diagnosis systems and information retrieval engines calibrated specifically for COVID-19. These applications rely on the ability to measure *semantic textual similarity* (STS), making STS a fundamental task that can benefit several downstream applications. However, existing STS datasets and models fail to translate their performance to a domain-specific environment such as COVID-19. To overcome this gap, we introduce CORD19STS dataset which includes 13,710 annotated sentence pairs collected from COVID-19 open research dataset (CORD-19) challenge. To be specific, we generated one million sentence pairs using different sampling strategies. We then used a finetuned BERT-like language model, which we call Sen-SCI-CORD19-BERT, to calculate the similarity scores between sentence pairs to provide a balanced dataset with respect to the different semantic similarity levels, which gives us a total of 32K sentence pairs. Each sentence pair was annotated by five Amazon Mechanical Turk (AMT) crowd workers, where the labels represent different semantic similarity levels between the sentence pairs (i.e. related, somewhat-related, and not-related). After employing a rigorous qualification tasks to verify collected annotations, our final CORD19STS dataset includes 13,710 sentence pairs.

## 1 Introduction

As a response to the worldwide COVID-19 pandemic, a massive amount of scientific literature related to coronavirus has emerged and continues to grow rapidly [1]. With the assistance of these documents, various natural language processing (NLP) applications can be developed to combat not only the ongoing COVID-19 pandemic but also to tackle the spread of infectious diseases in the future. Existing COVID-19 corpora can help conversational medical diagnosis systems [2], [19] to determine affected patients by matching textual symptom descriptions or signs to the most related pre-cached medical records. It enables intelligent information retrieval engines to filter out misleading, or malicious information spreading through social media [37], [43], [47]. It also benefits NLP applications on article categorizations [10], [27] into various sub-topics such as: disease precaution and disease evolution.

*The asterisk * next to author names denotes equal contribution.*

Measuring *sentence textual similarity* (STS) plays a key role in the above mentioned applications, and is well-explored mainly in NLP community [13], [35], [38], [39], [48]. These models are trained for STS over sentence pairs collected from non-scientific sources with annotations such as the ones summarized in Table 1. However, the use of these datasets for COVID-19-related NLP tasks might cause deteriorated performance due different data distribution. For example, STS benchmark (STSb) samples [14] could be extracted from general English language context, i.e. grounded image captions, indicating that it may fail to match the in-domain jargon for COVID-19 applications. Also, STS benchmarks proposed in the medical domain [44], [49] have scarce amount of labeled data, which limits their utility as a testbed for learning distributed representation. In this work, we introduce CORD19STS, a dataset for COVID-19 which consists of 13,710 annotated sentence pairs with semantic labels. The sentence pairs in our dataset were selected out of COVID-19 open research dataset (CORD-19) Challenge [1].

Specifically, the construction of CORD19STS contains data-collection and annotation. The data-collection is a two-stage process. First, we sample one million sentence pairs from CORD-19 by applying different sampling strategies to obtain diverse textual similarity pairs. Then, we employ a *balance selection scheme* to sample 32K sentence pairs out of the previous step, with the aim of balancing samples across different semantic similarity score intervals. In fact, our balance selection scheme is based on the computed similarity scores of the sentence pairs resulting from Sen-SCI-CORD19-BERT, which is the language model that combines many state-of-the-art efforts [1], [11], [40] for discerning textual semantic similarity between sentence pairs extracted from CORD-19. Data annotation, is done by qualified Amazon Mechanical Turk (AMT) crowd workers who pass pre- and post- qualification tests. We present detailed statistical analysis of our dataset.

## 2 Related work

**Sentence Textual Similarity** The standard NLP approach to obtain meaningful semantic sentence-level features is to learn sentence-level embeddings for the general purpose in the first place, then apply the model to a downstream, low-complexity task, such STS. Specifically, pretrained sentence level embeddings are finetuned over STS dataset for transferring generic sentence features towards ones with specific textual semantic meaning [35], [48], [51].

TABLE 1
Sentence pair datasets with their corresponding manually prepared STS annotations.

| Dataset | #Pairs | Domain | Annotators | Labels |
| --- | --- | --- | --- | --- |
| BIOSSES [44] | 100 | Biomedical | Medical experts | Similarity scores from 0 (no relation) to 4 (equivalent). |
| MedSTS [49] | 1,068 | Clinical | Medical experts | Semantic similarity scores from 0 to 5 (low to high similarity) |
| MedNLI [41] | 14,049 | Clinical | Radiologists | Entailment, Contradicts, Neutral |
| SNLI [12] | 570,000 | Grounded Human Writing | AMT users | Entailment, Contradicts, Neutral |
| STSb [14] | 8628 | Misc | AMT users | Similarity scores from 0 to 5 |
| CORD19STS | 13,710 | COVID-19 | AMT users | Related, Somewhat-related, Not-related |

The first step in STS is learning the distributed sentence representation with fixed-length, and this NLP task has been well-explored by many prior works. Unsupervised learning first has been commonly used for learning the sentence-level embedding [26], [28], however it has been reported in [17] that supervised sentence encoding method achieve the better performance over multiple transfer tasks. Also, two sentence encoding methods proposed in [15], the performance of which has been enhanced when supervised data is used to augment model training. It is worth mentioning that SNLI dataset [12], which offers a large-scale, naturalistic corpus of sentence pairs labeled for three different semantic scenarios between sentence pairs, is always used as the supervised signal for learning effective sentence representations [15], [17], [39]. Apart from that, most recently, many works [25], [34], [46], [48] demonstrate large-scale, multi-task learning helps model sentence-level embedding to gain generality and not be restricted to the single task and maintain high-standard performance over out-of-domain data.

However, comparing to the development of distributed sentence-level embedding learning algorithms, relatively little attention has been made to bring the new knowledge base for the STS task. Currently, STSb [13] is the standard benchmark [48], which has been extracted from corpus of STS shared task (2012-2017) [4], [5], [6], [7], [8], consisting around 8K samples collected from a general context corpus [13], which we argue may fail to offer insights for learning knowledge from the COVID-19 domain. Moreover, two medical STS datasets have been introduced by [44], [49], covering medical jargons and clinical notes, yet these two datasets do not offer enough amount of labeled data (around 1K at most as Table 1), and such nature of low-resource may be unable to serve as the benchmark for retrieving an effective distributed sentence features. Hence, we intend to share with the community CORD19STS, consisting of 13,710 annotated samples, with a target to offering a new dataset for learning sentence semantics for COVID-19.

**COVID-19 Related Research** Very recently, many research efforts have been made as the response to the unprecedented COVID-19 epidemic. Chen et al. [16] provide multilingual COVID-19 twitter dataset to track COVID-19-related misinformation and unverified rumors and study the public attitude and reaction on COVID-19 epidemic; this data collection is ongoing in the foreseeable future and has inspired many other works. The social media patterns are used for COVID-19 outbreak alignment across countries, modeling and predicting the pandemic spread [32]. The dataset of emotional responses to COVID-19 in text form has been introduce by [29], which consists of 5,000 annotated texts with corresponding emotion labels. A predictive modeling approach is employed to approximate the emotional responses of the texts. Spangher et al. in [45] have laid the methodological groundwork for the study of COVID policy, which benefits policy maker in the face of such pandemic crisis. In fact, not only have information on Tweet been studied in different languages [9], [31], but COVID-19 related messages on Reddit and YouTube are also involved in the recent study [3], [42].

Apart from the social-media data collection, a coalition of leading research groups have prepared the COVID-19 Open Research Dataset (CORD-19) [1], which gathers papers and preprints historically and presently pertinent to COVID-19 from Semantic Scholar. CORD-19 serves as the sound research foundation for many following works including ours. Huang et al. [21] collect segmented section annotation over 10,966 abstracts of CORD19 dataset, in which annotations provided by AMT users are highly-aligned with that of experts. COVID-Q [22] present annotations for 1,690 questions out of CORD19 dataset, where the labels determine the type of the questions asked within each text. They categorize questions to 15 classes and leverage BERT as their baseline. Moller et al. [23] collect 2,019 question-answer pairs annotated by volunteer biomedical experts. Both two works [22], [23] shows improved performance in QA related task. It is worth to mention that our CORD19STS dataset mainly focuses on offering information over sentence-level semantic meaning, which is orthogonal to the prior works in [21], [22], [23], [24].

## 3 PRELIMINARY ON CORD-19 DATASET

In this section, we provide a brief review of CORD-19 [1], which plays a significant role in the CORD19STS construction.

CORD-19 Dataset is collected out of 128,000 scholarly articles including over 59,000 full texts on COVID-19, SARS-CoV-2, and related coronaviruses. It integrates papers from several openly accessible sources, with the aim of bridging different research communities around the same scientific cause. The dataset can be downloaded from kaggle website and it contains JSON files of the full text of the PDFs in a structured format. General statistics of CORD19 dataset such as number of documents per category, and number of sentences are provided in Table 2.

TABLE 2
Statistics on CORD19 dataset.

| Corpus | Documents | Sentences |
|---|---|---|
| Comm-use [1] | 9,557 | 319,299 |
| Non-comm [1] | 2,466 | 453,610 |
| Custom [1] | 27,220 | 5,984,201 |
| Biorxiv [1] | 1,934 | 73,274 |
| Overall [1] | 41,177 | 6,830,384 |

Given the collected abstract and body-texts of the JSON files of CORD-19 dataset, we applied different preprocessing steps including lower casing the text, removing special characters, citation removal, and sentence tokenization.

## 4 CONSTRUCTION OF CORD19STS

In this section, we provide detailed information of our data-collection process including: selection of one million sentence pairs by appying *four different sentence pair sampling strategies* (Section 4.1); our proposed language model Sen-SCI-CORD19-BERT to compute similarity scores of the selected one million sentence pairs (Section 4.2); our applied *balance sampling scheme* to filter out sentence pairs, for which their computed similarity scores resulting from Sen-SCI-CORD19-BERT model are uniformly distributed within different interval scores (Section 4.3); and details of our AMT setting to collect data annotations by AMT workers (Section 4.4).

### 4.1 Sentence-Pair Sampling Strategies

We employ four different sampling strategies in our pipeline, in order to generate sentence pairs with diverse underlying semantic similarity patterns, i.e. covering different levels of semantic similarities that can be inherently related to the text spanning distance, as follows: (S1) selecting two consecutive sentences within a same paragraph of a document, (S2) selecting pairs of sentences from same paragraphs, (S3) selecting sentence pairs from different paragraphs of a same document, and (S4) selecting sentence pairs from two different documents.

Examples of the four different strategies (S1-S4) are shown in Fig. 1, in which we can observe a meaningful correlation between the physical text spanning distance and semantic similarity level. The anchor sentence contains *"gene pair might have already been there"* (in orange color) can be expressed as *"gene pairs that we can observe"*, contained in the sentence sampled by S1. Such correlation gradually degrades over sentences collected by S2 and S3, which merely contains overlapping words such as *"gene pair"* and *"ancestral"* but hardly express the meaning of the anchor sentence. Finally, it can be seen that the sentence sampled by S4 does not have much semantic correlation with the anchor sentence.

Applying the above mentioned sampling strategies, we collect one million sentence pairs from CORD-19. Although, the initial intuition of applying them was to create sentence pairs reflecting different levels of similarities, i.e. having high to low levels of sentence pair similarities while moving from strategy S1 to S4, surprisingly there still exist many sentence pairs with low level of similarities comparing to high level ones even applying the first two strategies (S1 and S2).

### 4.2 STS Computation using Our Proposed Sen-SCI-CORD19-BERT Model

Sen-SCI-CORD19-BERT is our proposed language model for measuring textual similarity between sentence pairs of STSCORD19 dataset. To be specific, SCIBERT [11] is a pretrained BERT model over 1.14M papers from semantic scholar, and proves its effectiveness over a suite of tasks and datasets from scientific domains, such as Named Entity Recognition (NER) and Relation Classification (REL) [20], [30], [33]. Due to the fact that CORD-19 has been established over a large amount of scientific articles and pre-prints from semantic scholar, we employ SCI-BERT for modeling the textual information. More formally, we fine-tune SCI-BERT over preprocessed CORD-19 text via Masked Language Modeling following [18]. We call the resulting language model SCI-CORD19-BERT.

Meanwhile, Sentence-BERT (SBERT) [40] is the state-of-the-art method for efficiently retrieving sentence pairs with textual semantic similarity, measured as the cosine distance between the two fixed-dimension sentence embeddings. In fact, SBERT incorporates a pretrained language model followed by a pooling mechanism, into a Siamese network architecture to compute the similarity scores, and the training procedure is orthogonal to the word-level language models such as BERT [18] and RoBERTa [36]. Therefore, we combine the SCI-CORD19-BERT with the Siamese architecture as Sen-SCI-CORD19-BERT, for obtaining the semantic similarity level for CORD-19 sentence pairs, as shown in Figure 2.

### 4.3 Sampling-Balance Scheme

Given the one million sentence pairs generated out of the prior step, we attempt to balance them with respect to the different semantic similarity levels. To do so, we use our proposed Sen-SCI-CORD19-BERT model (Section 3) to compute similarity scores of the one million samples. We set different score intervals as $\{(-1, 0.25), (0.25, 0.5), (0.5, 0.75), (0.75, 1)\}$. For each interval, we sample 2000 sentence pairs per sampling strategy (8000 = 2000*4 samples per interval). As a result, we collect 32K (=4*8K) sentences in total and largely maintain the balance in terms of both underlying similarity patterns and computed similarity levels.

Distributions of the computed similarity scores of the one million sentence pairs are shown in Fig. 3. We address the problem of having unbalanced samples over different interval scores by applying the balance selection scheme, which helps to obtain a dataset balanced over similarity score levels as well as the strategies.

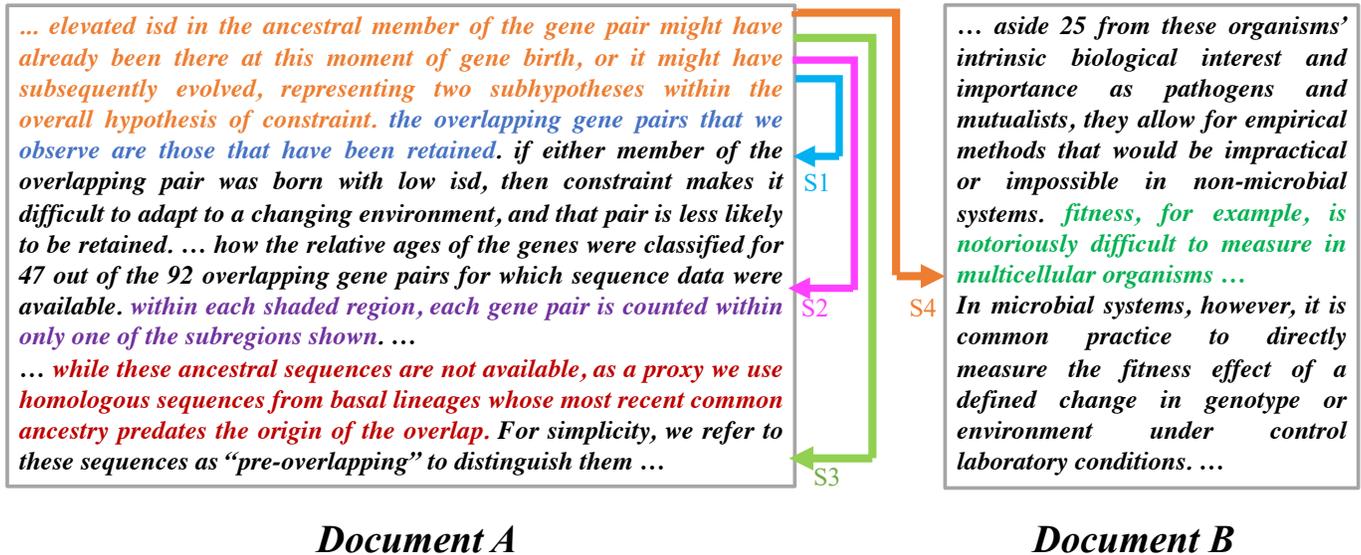

Fig. 1. Visualization of our four different sentence pair sampling strategies: *consecutive sentences* (S1), *sentences from a same paragraph* (S2), *sentences from a same document* (S3) and *sentences from different documents* (S4).

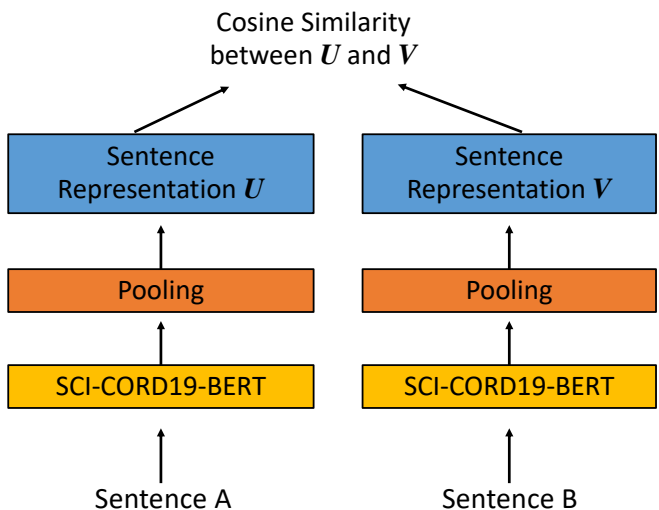

Fig. 2. The visualization over Sen-SCI-CORD19-BERT, which incorporates SCI-CORD19-BERT with the pooling mechanism into a siamese network.

### 4.4 Amazon Mechanical Turk Annotation

Annotation of the sentence pairs of our dataset were provided by AMT crowd users. In our AMT API, the labels were set to three different levels: {related, somewhat-related, not-related}. To reduce potential subjectwise bias introduced to the annotations, each individual sentence pair was assigned to five different annotators. **Pre-Qualification:** Our AMT pipeline had a pre-test qualification part, by which, we first train AMT users by providing them examples of sentence pairs with their correct corresponding labels (Table 5), and then, ask them to attend to our pretest quiz including six sentence pairs. Users providing at least four correct labelings were allowed to attend to our data annotation.

**Post-qualification:** We also include some hidden qualification test samples at each AMT-HIT to verify the performance of the users. Very poor performance on those hidden samples leads to AMT-HIT rejection. We have to take this step because some annotators have tried to game the system by annotating without thinking leading to useless annotations.

Fig. 4 illustrates the statistics of the AMT workers in terms of worker performance over the hidden qualification questions (a-axis) versus total number of the HITs made by each worker (y-axis) and the average time spent per HIT by the worker, which is encoded/proportional to the radii of the scatter plot. As a data cleaning step in our pipeline, we exclude HITs for which AMT users did not have a good performance over the hidden questions or the ones who did not spend enough time per HIT.

**Setting the gold labels** After verifying AMT users following the above mentioned criteria, we set the gold label of the pairs as the label with a minimum of three concurring labels.

### 5 STATISTICS OF THE COLLECTED ANNOTATIONS

Fig. 5 visualizes statistics of the final gold labels per strategy (Figures 5a-5d) as well as overall (Figure 5e). The NULL bar represents numbers of the sentence pairs which did not have at least three concurring annotations. Distribution of the textual similarity scores over different labels are shown in Figures 5f-5j.

### 5.1 Sen-SCI-CORD19-BERT Evaluation

Since we design our sampling method based on Sen-SCI-CORD19-BERT, it is important to validate its performance for modeling language. In this work, we evaluate Sen-SCI-CORD19-BERT based on two experiments, more details of which can be found at our project page[1]. **Performance on STSb dataset.** STSb dataset contains 8628 sentence pairs,

---

1. https://gitlab.vista.isi.edu/xiaoguo/cord_19/tree/master

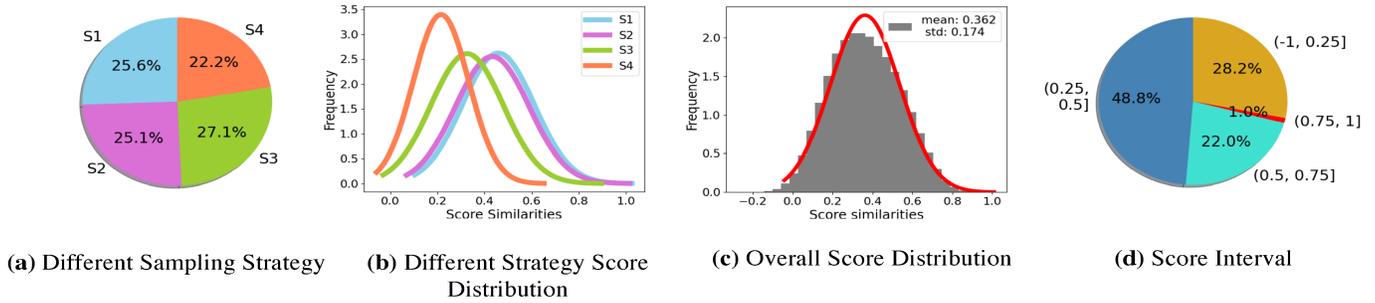

(a) Different Sampling Strategy  (b) Different Strategy Score Distribution  (c) Overall Score Distribution  (d) Score Interval

Fig. 3. Statistics of the different sentence pair strategies, The STS distribution scores computed as the cosine similarity between the sentence embeddings resulting from the proposed Sen-SCI-CORD19BERT are shown for: *consecutive sentences* (S1), *sentences from a same paragraph* (S2), *sentences from a same document* (S3) and *sentences from different documents* (S4). **(a).** Different percentage of sentence pairs collected by different sampling strategies, more details are reported in Table 4. **(b).** The overlaid Gaussian distribution of the different strategies. Except for (S1) and (S2), any given two distributions share with different features, indicating they own different latent similarity patterns. **(c).** The similarity score histogram and Gaussian distribution for overall samples. **(d).** The score interval of overall samples. It can be observed that samples with similarity scores within (0.75, 1] merely constitutes 1.0%, which motivates to propose *Sampling-Balance Scheme*.

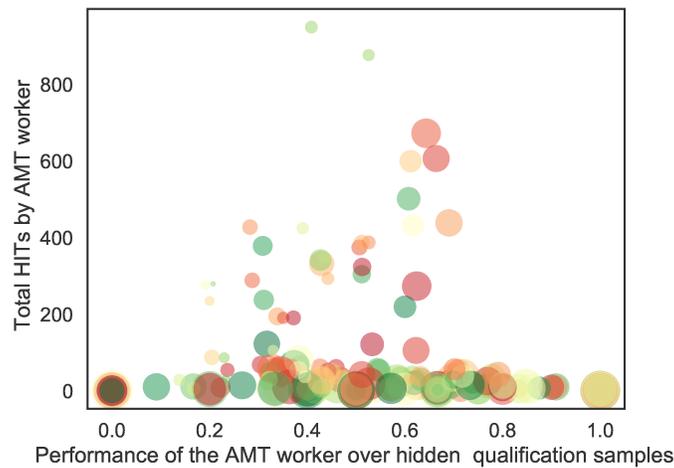

Fig. 4. Visualization of the statistics of the AMT workers in terms of: worker performance over the hidden qualification questions (a-axis); total number of the HITs made by each worker (y-axis); and the average time spent per HIT by the worker, which is encoded/proportional to the radii of the scatter plot. As a data cleaning step in our pipeline, we exclude HITs for which AMT users did not have a good performance over the hidden questions or the ones who did not spend enough time per HIT.

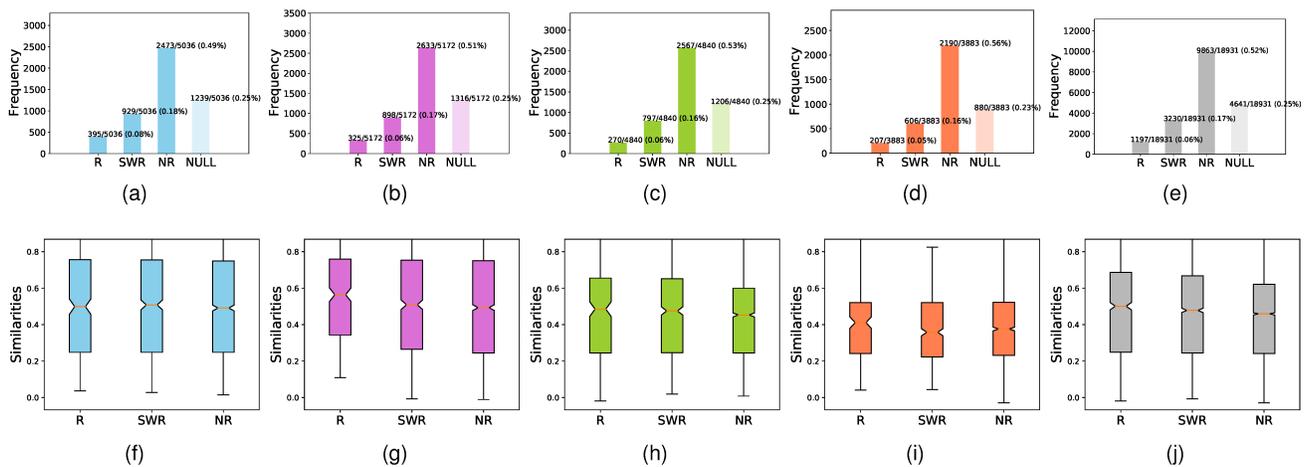

Fig. 5. Statistics of the final gold labels (R: related, SWR: somewhat-related, NR: not-related) per strategy (a-d) as well as overall (e). The NULL bar represents numbers of the sentence pairs which did not have at least three concurring annotations. Distribution of the textual similarity scores over different labels are shown (f-j).

and we follow the data partition defined in [13] which train, validation and test sets have 5749, 1500, 1379 sentence pairs respectively. In order to report a comprehensive model performance, we conduct experiment over supervised and unsupervised settings, following [40]. Specifically, in the unsupervised manner, models are trained over NLI dataset, the combination of SNLI [12] and Multi-Genre NLI [50]; and in the supervised manner, models are training on either the NLI first then STSb dataset, or only STSb dataset.

We report Spearman correlation in the Table 3. Our proposed Sen-SCI-CORD19-BERT and Sen-SCI-BERT can produce meaningful semantic similarity results over sentences extracted in the general background. However, finetuning over CORD19 can slightly reduce the performance, which explains why they are slightly worse than models proposed in the previous work.

| Model | Spearman |
|---|---|
| *Unsupervised (trained on NLI data)* | |
| InferSent - GloVe [15] | 68.03 |
| Universal Sentence Encoder [17] | 74.92 |
| SBERT | 77.03 |
| Sen-SCI-BERT | 75.32 |
| Sen-SCI-CORD19-BERT | 75.09 |
| *Supervised (trained on STSb data)* | |
| SBERT | 84.67 ± 0.19 |
| SRoBERTa | 84.92 ± 0.34 |
| Sen-SCI-BERT | 80.44 ± 0.19 |
| Sen-SCI-CORD19-BERT | 80.60 ± 0.22 |
| *Supervised (trained on NLI + STSb data)* | |
| SBERT | 85.35 ± 0.17 |
| SRoBERTa | 84.79 ± 0.38 |
| Sen-SCI-BERT | 82.22 ± 0.05 |
| Sen-SCI-CORD19-BERT | 82.91 ± 0.05 |

TABLE 3
The comparison on supervised and unsupervised STS.

## 6 CONCLUSIONS

In this work, we propose CORD19STS, a semantic textual similarity dataset for COVID-19, which solve the issue that previous STS dataset lacks the in-domain knowledge for the outbreak worldwide pandemic. In order to obtain a dataset which is balanced in terms of latent similarity patterns and different similarity levels, we use four different sentence-pair sampling strategies with balance sampling scheme that relies on our proposed language model, Sen-SCI-CORD19-BERT. Also, we design multiple qualification for AMT to guarantee high-quality annotation can be provided. To this end, we present detailed statistical analysis on the proposed CORD19STS, and evaluate our proposed language model.

# 7 APPENDIX

| Sampling Strategy | S1 | S2 | S3 | S4 |
|---|---|---|---|---|
| Comm-use | 59,737 | 81,440 | 80,636 | 47,507 |
| Non-comm | 13,131 | 9,932 | 14,122 | 13,648 |
| Custom | 168,158 | 146,316 | 158,352 | 148,419 |
| Biorxiv | 7,163 | 6,272 | 9,597 | 6,168 |
| Overall | 248,209 | 244,011 | 262,707 | 215,742 |

TABLE 4
Samples number obtained by different sampling strategies over different sub-dataset.

| Ex. | Sentence-Pair | Label |
| --- | --- | --- |
| #1.a | **Text#1**: researchers believe that americans with diabetes , chronic lung disease , and cardiovascular disease all diseases linked to obesity are at higher cov risk<br>**Text#2**: major risk factors for covid include obesity , diabetes , and cardiovascular diseases<br>**Explanation:** both sentences are talking about a set of similar risk factors on covid | Related |
| #1.b | **Text#1**: researchers believe that americans with diabetes , chronic lung disease , and cardiovascular disease all diseases linked to obesity are at higher cov risk<br>**Text#2**: 30 minutes of moderate-intensity aerobic activity like dog walking helps lower blood pressure and improve overall cardiovascular health.<br>**Explanation:** both sentences are related to cardiovascular diseases but not addressing thesame topic | Somewhat-related |
| #1.c | **Text#1**: researchers believe that americans with diabetes , chronic lung disease , and cardiovascular disease all diseases linked to obesity are at higher cov risk<br>**Text#2**: this structural and immune characterization provides new insights into coronavirus spike stability determinants and elores the immune landscape of viral spike proteins .<br>**Explanation:** it is not easy to find any mutual connection between the two sentences | Not-related |
| #2.a | **Text#1**: covid - 19 does not care about your race , ethnicity , gender , age , or immigration status<br>**Text#2**: it has been shown that demographic characteristics might have some correlation with sars<br>**Explanation:** both sentences are talking about the effect of demographic features such as: race , ethnicity , gender, and age on covid/sars | Related |
| #2.b | **Text#1**: covid - 19 does not care about your race , ethnicity , gender , age , or immigration status<br>**Text#2**: Information was collected on potential risk factors for SARS-CoV infection (such as having a chronic disease), personal hygiene (such as washing hands), and the use of masks<br>**Explanation:** sars/covid risk factors from completely two different categories are mentioned within the two sentences | Somewhat-related |
| #2.c | **Text#1**: covid - 19 does not care about your race , ethnicity , gender , age , or immigration status<br>**Text#2**: we apply an intervention analysis to assess if this influenza season deviates from eectations<br>**Explanation:** although the second sentence is related to influenza, which is a similar disease to covid, but it has nothing to do with the risk factors | Not-related |
| #3.a | **Text#1**: the mask also protects you against an infected person sneezing and coughing airborne covid and prevent it from entering your nose or mouth<br>**Text#2**: individual level studies have found that the use of face masks was protective for the acquisition and transmission of a range of respiratory viruses including sars cov - 1<br>**Explanation:** both sentences are talking about the effectiveness of wearing mask on sars/covid disease | Related |
| #3.b | **Text#1**: the mask also protects you against an infected person sneezing and coughing airborne covid and prevent it from entering your nose or mouth<br>**Text#2**: disposable medical face masks are intended for a single use only<br>**Explanation:** both sentences are related to mask; but one is on its effectiveness and the other one is on the importance of changing disposable masks | Somewhat-related |
| #3.c | **Text#1**: the mask also protects you against an infected person sneezing and coughing airborne covid and prevent it from entering your nose or mouth<br>**Text#2**: linear mixed models revealed that the presence of mandated bcg policies was associated with a significant flattening of the enential increase in both confirmed cases and deaths during the first - day period of country - wise outbreaks .<br>**Explanation:** not easy to find any commonality between the two sentences | Not-related |
| #4.a | **Text#1**: we aimed to develop a deep learning method that could extract covid - 19 's graphical features in order to provide a clinical diagnosis ahead of the pathogenic test , thus saving critical time for disease control<br>**Text#2**: we established an auxiliary diagnostic tool based on artificial intelligent algorithm to diagnostic hyperlipemia and automatically predict the corresponding diagnostic markers using hematological parameters<br>**Explanation:** both sentences are talking about applying machine learning techniques for disease predictions | Related |
| #4.b | **Text#1**: we aimed to develop a deep learning method that could extract covid - 19 's graphical features in order to provide a clinical diagnosis ahead of the pathogenic test , thus saving critical time for disease control<br>**Text#2**: our method is based on naive bayes, a machine learning algorithm which uses the observed frequencies in the training dataset to estimate the probability that a pair is linked given a set of covariates<br>**Explanation:** two sentences have commonalities on applying machine learning | Somewhat-related |
| #4.c | **Text#1**: we aimed to develop a deep learning method that could extract covid - 19 's graphical features in order to provide a clinical diagnosis ahead of the pathogenic test , thus saving critical time for disease control<br>**Text#2**: we characterized importations timeline to assess the rapidity of isolation , and epidemiologically linked clusters to estimate the rate of detection<br>**Explanation:** not easy to find any commonality between the two sentences | Not-related |

TABLE 5
Sample sentence pairs with their corresponding semantic similarity scores/labels provided to the AMT users in the introduction section.